\documentclass[10pt,twocolumn,letterpaper]{article}

\usepackage{main}              

\usepackage{multirow}
\usepackage{tabularx}
\usepackage[dvipsnames]{xcolor}
\usepackage{lipsum}
\usepackage[linesnumbered,ruled,vlined]{algorithm2e}

\definecolor{cvprblue}{rgb}{0.21,0.49,0.74}
\usepackage[pagebackref,breaklinks,colorlinks,citecolor=cvprblue]{hyperref}

\title{SNP: Structured Neuron-level Pruning to Preserve Attention Scores}

\author{Kyunghwan~Shim$^{1}$ Jaewoong~Yun$^{1}$ Shinkook~Choi$^{1}$\\
Nota Inc.$^{1}$\\
{\tt\small \{kyunghwan.shim, jwyun, shinkook.choi\}@nota.ai}
}

\begin{document}
\maketitle

\begin{abstract}

Multi-head self-attention (MSA) is a key component of Vision Transformers (ViTs), which have achieved great success in various vision tasks. However, their high computational cost and memory footprint hinder their deployment on resource-constrained devices. 
Conventional pruning approaches can only compress and accelerate the MSA module using head pruning, although the head is not an atomic unit. 
To address this issue, we propose a novel graph-aware neuron-level pruning method, Structured Neuron-level Pruning (SNP). SNP prunes neurons with less informative attention scores and eliminates redundancy among heads. 
Specifically, it prunes graphically connected query and key layers having the least informative attention scores while preserving the overall attention scores. Value layers, which can be pruned independently, are pruned to eliminate inter-head redundancy. Our proposed method effectively compresses and accelerates Transformer-based models for both edge devices and server processors. 
For instance, the DeiT-Small with SNP runs 3.1$\times$ faster than the original model and achieves performance that is 21.94\% faster and 1.12\% higher than the DeiT-Tiny. 
Additionally, SNP combine successfully with conventional head or block pruning approaches. 
SNP with head pruning could compress the DeiT-Base by 80\% of the parameters and computational costs and achieve 3.85$\times$ faster inference speed on RTX3090 and 4.93$\times$ on Jetson Nano.
\end{abstract} 
\begin{figure}
    \centering
    \includegraphics[width=1.0\columnwidth]{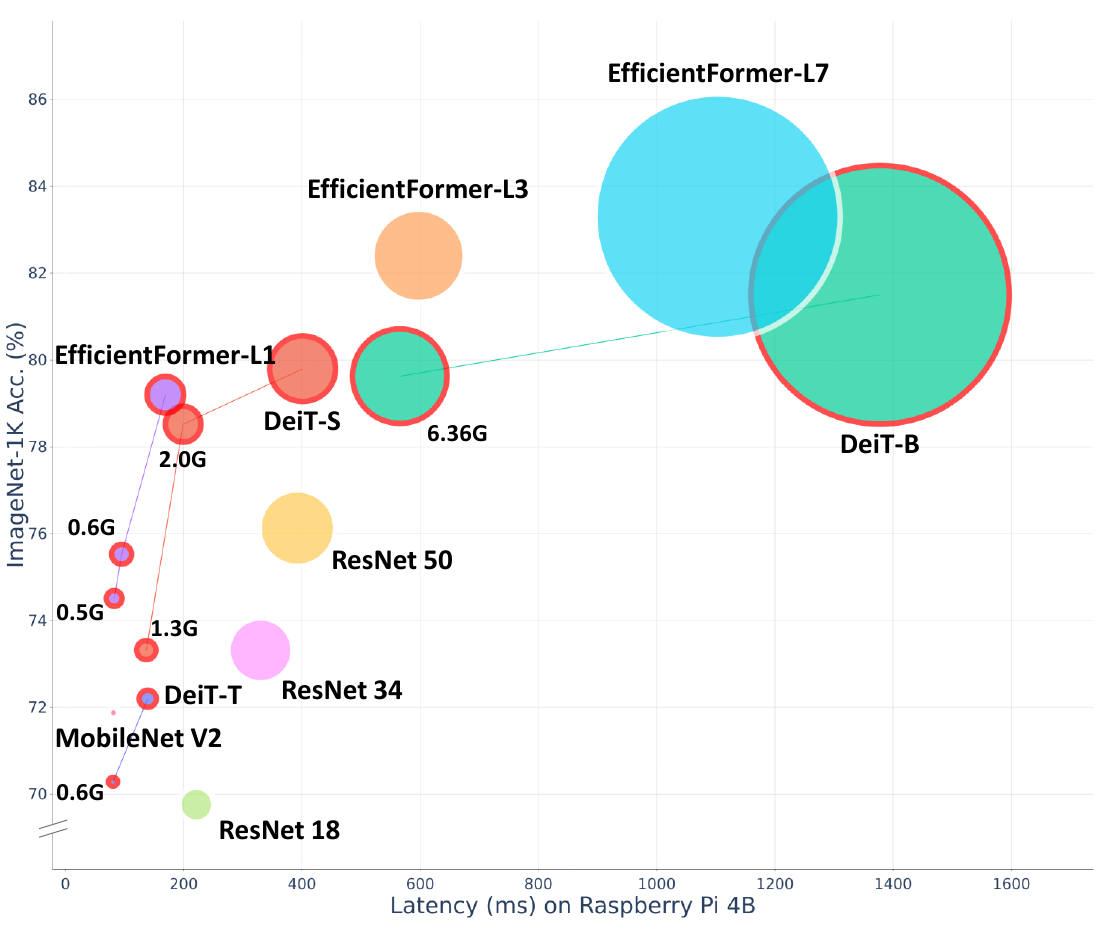}
    \caption{\textbf{Comparison of model size, speed, and performance.} ImageNet-$1\mathrm{K}$ classification results. Latency is profiled by Rasbperry Pi 4B. The connected lines represent the compressed models paired with the original model. The size of each circle indicates the number of parameters in respective model. The number adjacent to each compressed model indicates its compressed GFLOPs.}
    \label{overall_perf}
\end{figure}

\section{Introduction}
\label{sec:intro}

Vision Transformers (ViTs)~\cite{dosovitskiy2020image, touvron2021training_deit, liu2021swin} have outperformed or matched the performance of state-of-the-art convolutional neural networks (CNNs)~\cite{he2016deep, tan2019efficientnet, radosavovic2020designing_regnet, xie2020self} on various computer vision tasks. The success of ViTs is attributed to the Multi-head Self-Attention (MSA) module, which captures intricate relationships in data. However, the Transformer architecture entails substantial computational resources, posing challenges for practical applications on edge devices with constrained storage and computational capabilities. To address this issue, we leverage the graphical components of the MSA module to reduce the dimension of interconnected layers, aimed at effectively reducing their computing budgets while achieving hardware-agnostic speedups.

From the perspective of structured pruning, which aims to reduce the number of dimensions in convolutional or linear layers, the MSA module contains two prunable objectives: heads and neurons. 
Head pruning, which removes the number of heads, is relatively intuitive to implement due to the reduced complexity of graphical elements compared to neuron-level pruning.
In contrast, neuron-level pruning reduces the dimension of individual layers in each head of the MSA module, necessitating a comprehensive understanding of the graphical connectivity of the MSA module. 
A recent work~\cite{uvc} implements neuron-level pruning by zeroing out individual filters without considering the graphical connectivity of the model. This indiscriminate zeroing can negatively affect both accuracy and throughput performance~\cite{wan2023upscale}.

In this paper, we introduce a novel graph-aware neuron-level pruning method called Structured Neuron-level Pruning (SNP) to accelerate and compress ViTs effectively. 
We propose two pruning criteria based on the function of each layer within the MSA module.
SNP prunes filter pairs of query and key layers containing fewer contributions to attention scores.
Moreover, SNP aims to reduce redundancy across heads by eliminating the redundant filter from the value layer.

Furthermore, by removing identical filter indices across all graphically connected layers, SNP can accelerate various Transformer models on various devices without additional libraries, as shown in \cref{overall_perf}.

In summary, the major contributions of this paper are:
\begin{itemize}
    \item We propose a novel graph-aware neuron-level pruning method, SNP, for Transformer models. SNP is the first method to use the graphical characteristics of the MSA module to measure the importance score of neurons.
    \item To the best of our knowledge, this is the first work to accelerate Transformer models using neuron-level pruning only.
    \item SNP achieves significant acceleration while maintaining the original performance on several models. Compressed DeiT-Small outperforms DeiT-Tiny by 1.12\% in accuracy, with similar FLOPs, and reduces inference time on various edge devices. Additionally, the proposed method accelerates the efficiently designed Transformer model, EfficientFormer~\cite{li2022efficientformer}, more than two times with acceptable performance degradation.
\end{itemize}

\begin{figure*}
    \centering
    \includegraphics[scale=0.9]{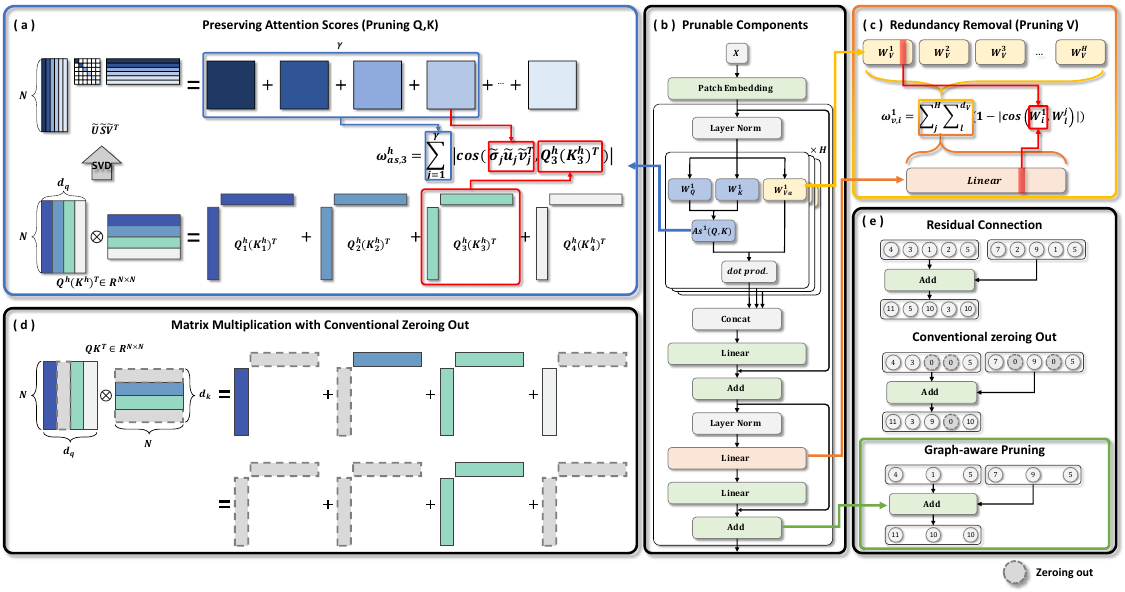}
    \caption{
    \textbf{Proposed SNP methods, on each prunable component of the Transformer block.}
    \textbf{(a)} SNP pruning criteria of query and key layers to preserve attention scores. 
    \textbf{(b)} prunable components of Transformer block. 
    \textbf{(c)} SNP pruning criteria of value and other layers, including FFN and patch embedding.
    \textbf{(d)} conventional zeroing out in the matrix multiplication operator. 
    \textbf{(e)} Conventional zeroing out and graph-aware pruning in the residual connection.
    }
    \label{fig:proposed_methods}
\end{figure*}


\section{Related Work}
\subsection{Compressing vision Transformers}

ViTs~\cite{dosovitskiy2020image, touvron2021training_deit, liu2021swin} achieve high performance in numerous vision tasks without specialized image processing modules such as convolutions. The key concept of ViTs is to segment images into patch sequences, convert these patches to token embeddings, and then process them through Transformer encoders~\cite{vaswani2017attention}. 
ViTs consist only of Transformer blocks, making them likely to be over-parameterized. For this reason, recent works have aimed to reduce computational cost~\cite{touvron2021training_deit, bolya2022token} and be memory efficient~\cite{liu2021post, wan2023upscale}. 
DeiT~\cite{touvron2021training_deit} proposes lightweight ViT architectures through knowledge distillation~\cite{hinton2015distilling}. ToMe~\cite{bolya2022token} proposes accelerating ViTs by directly combining similar tokens, without the need for training. Liu~\etal~\cite{liu2021post} propose a post-training quantization method using a mixed precision scheme based on nuclear norm that does not require fine-tuning for the vision Transformer.

\subsection{Pruning vision Transformers}

\subsubsection{Unstructured and structured pruning}
Pruning methods can be broadly categorized into two types, unstructured and structured pruning. 
Unstructured pruning sets individual weights or parameters to zero, resulting in irregular sparse matrices~\cite{han2015deep, lee2020layer}. Compressed models using unstructured pruning tend to maintain relatively high performance for a given pruning ratio. However, they necessitate additional libraries, such as cuSPARSE~\cite{cusparse}, Automatic SParsity~\cite{mishra2021accelerating}, or SparseDNN~\cite{wang2021sparsednn} to accelerate sparse matrix computations.

Structured pruning, on the other hand, involves the removal of entire groups of units, such as filters or attention heads. This can be implemented using “masking" (zeroing out)~\cite{yu_mask_reinforcement, yu_brain, gm_mask_0, He_channel_masking}, or by “pruning"~\cite{li2016pruning_first_filter_pruning, vainf_pruning}. Structured pruning by masking~\cite{yu_mask_reinforcement, yu_brain, gm_mask_0, He_channel_masking} simply sets the group of units to zero, which requires additional libraries to accelerate the model, as unstructured pruning. “Pruning"~\cite{li2016pruning_first_filter_pruning,vainf_pruning}, on the other hand, requires a comprehensive understanding of the network's graphical connectivity, including element-wise operations that enforce the same input shape. By considering the graphical connectivity and pruning identical filter indices for interconnected layers, structured pruning can achieve acceleration on any devices.

\subsubsection{Head and neuron-level pruning}
Structured pruning for the MSA module has two pruning objectives: head and neuron. Head pruning~\cite{xpruner, WDPruning} reduces the number of heads, while neuron-level pruning~\cite{uvc} reduces the dimension of each query, key, and value layer in each head. Recent studies for pruning ViTs have focused on head pruning. X-Pruner~\cite{xpruner} proposes a novel head pruning method for ViTs that introduces explainability-aware masks and measures the importance of the head, resulting in superior model compression. WDPruning~\cite{WDPruning} proposes a method to control the number of attention heads and blocks via threshold on learnable parameters.

UVC~\cite{uvc} utilizes knowledge distillation alongside several pruning techniques, such as head pruning, block pruning, and neuron-level pruning. 
However, the neuron-level pruning of UVC is carried out in a masking (zeroing out) manner, converting the weight matrix into a sparse matrix.
For this reason, UVC necessitates additional libraries or hardware for accelerating sparse matrices, otherwise, the compressed model with UVC cannot achieve latency gain from the neuron-level pruning.
\section{Methodology}


\subsection{Preliminaries}

MSA module takes a single input $X\in \mathbb{R}^{N\times d}$, where $N$ denotes the input vector length, and $d$ represents the hidden size. It comprises $H$ heads, each consisting of three linear layers: query, key, and value. Each layer denoted as $W_{\{q,k,v\}}^{h} \in \mathbb{R}^{d\times d_{\{q,k,v\}}}$, where $h$ represents the $h$-th head and $d_{\{q,k,v\}}$ indicates the hidden size of each query, key, and value layer. ${\{Q,K,V\}}^{h}$ signifies the output of each query, key, and value layer in the $h$ th head, with a shape of ${\mathbb{R}^{N\times d_{\{q,k,v\}}}}$.

The self-attention operation for $h$-th head can be expressed as follows:
\begin{equation}
\label{eq:attention_score}
    \mathrm{As}^{h}(X)= Q^{h} \cdot (K^{h})^{T} \\
\end{equation}
\begin{equation}
\label{eq:attention_module}
    \mathrm{Att}^{h}(X) = \mathrm{Softmax}(\frac{\mathrm{As}^{h}(X)}{\sqrt{d_{q}}}) \cdot V^{h}
\end{equation}
\begin{equation}
    \label{concat_msa_module}
    \mathrm{MSA}(X) = \mathrm{Concat}(\mathrm{Att}^{1}(X), ..., \mathrm{Att}^{H}(X))
\end{equation}
here, $\mathrm{As}$, $\mathrm{Att}$, and $\mathrm{MSA}$ represent the functions to calculate attention scores, attention module, and MSA module respectively.

Matrix multiplication, unlike a residual connection, yields zero when either of the inputs is masked, regardless of the value of the other input. Therefore, applying a conventional zeroing-out approach to neuron-level pruning in the MSA module can lead to unexpected results, as shown in \cref{fig:proposed_methods} (d). To address this issue, we introduce two graph-aware neuron-level pruning criteria to compress and expedite the MSA module.

\subsection{Preserving attention scores}
\label{sec:preserving_attention_scores}

Attention scores~\cref{eq:attention_score} of the MSA module learn long-range dependencies between image features by focusing on different parts of the image when processing different features.
These attention scores can be recognized as a series of outer product of $Q^{h}_{i}$ and $(K^{h}_{i})^T$ as follows: 
\begin{equation}
    \label{eq:QK_w_outerproduct}
    \begin{aligned}
        \mathrm{As}^{h}(X) 
        &= Q^{h} \cdot (K^{h})^{T} \\
        &= \sum^{d_{q}}_{i=1}{Q^{h}_{i} \cdot (K^{h}_{i})^{T}}\\
    \end{aligned}
\end{equation}
from this perspective, neuron-level pruning, reducing the dimension of query $d_{q}$ and key $d_{k}$, inevitably distorts the attention scores.
For this reason, preserving the attention scores is essential to maintain the high performance of the original model. 

To alleviate the distortion, we maintain the graphically connected query-key filter pair ($Q^{h}_{i}$ and $K^{h}_{i}$), constituting a filter-by-attention score ($Q^{h}_{i}\cdot (K^{h}_{i})^{T} \in \mathbb{R}^{N\times N}$), that retains the most significant aspects of the overall attention scores.
To identify the most informative filter pair, we initially employ singular value decomposition (SVD) to decompose the original model's attention scores.
\begin{equation}
    \begin{aligned}
        \mathrm{As}^{h}(X) 
        &= \tilde{U} \cdot \tilde{S} \cdot \tilde{V}^{T} \\
        &= \sum^{N}_{j=1}{\tilde{\sigma}_{j} \cdot \tilde{u}_{j}\cdot \tilde{v}_{j}^{T}} \\
    \end{aligned}
    \label{eq:SVD_on_attention_score}
\end{equation}
where $\tilde{U}$ and $\tilde{V}$ are the left and right singular vector matrices, respectively, and $\tilde{S}$ is the diagonal matrix of singular values, with $\tilde{\sigma}_{1} \geq \tilde{\sigma}_{2} \geq...\geq \tilde{\sigma}_{N}$. 

SVD is a technique for extracting the most informative components of a matrix, those with large singular values, while discarding the less informative components, those with small singular values. 
To retain the spatial relationships captured by the attention mechanism, we prune the filter pair $Q_{i}^{h}$ and $K_{i}^{h}$ with the least correlation with the most informative components of the attention scores.

To measure the correlation, we adopt the cosine similarity between the attention scores of the $i$-th filter and the $j$-th rank matrix. Consequently, the importance score $\omega^{h}_{as,i}$ is defined as follows:
\begin{equation}
    \begin{aligned}
    \omega_{as,i}^{h}
    &= \sum_{j=1}^{r}{|\mathrm{cos}((Q^{h}_{i}\cdot (K^{h}_{i})^{T}), (\tilde{\sigma}_{j}\cdot \tilde{u}_{j} \cdot \tilde{v}_{j}^{T}))|}\\
    &= \sum_{j=1}^{r}{|\frac{(Q^{h}_{i}\cdot (K^{h}_{i})^{T}) \cdot (\tilde{\sigma}_{j} \cdot \tilde{u}_{j} \cdot \tilde{v}_{j}^{T})}{||(Q^{h}_{i}\cdot (K^{h}_{i})^{T})|| \cdot ||(\tilde{\sigma}_{j} \cdot \tilde{u}_{j} \cdot \tilde{v}_{j}^{T})||}|}
    \end{aligned}
    \label{eq:cos_sim}
\end{equation}
where $r$ is a hyperparameter dictating the quantity of ranks, within the range of $1 \leq r\leq N$, to be compared with the $i$-th attention scores, while the remaining $N-r$ singular values ($\tilde{\sigma}_{r+1}, ..., \tilde{\sigma}_{N}$) are discarded. 

Optimizing $r$ for each attention module can contribute to preserving informative filters and sustaining higher performance. 
However, the optimal value of $r$ may vary according to the model's domain task or trained datasets, if $r$ is too low, important components of the attention scores can be removed, leading to a significant performance degradation.
Therefore, we set $r$ to its full rank $N$, even though it might be sub-optimal. 
Despite this, SNP demonstrates superior performance across various Transformer models.

The importance score $\omega^{h}_{as,i}$, defined in Eq.~(\ref{eq:cos_sim}), represents the importance of the $i$-th filter for both query and key layers. A larger $\omega^{h}_{as,i}$ indicates that the filter has a greater impact on the main component of the attention scores, while a lower $\omega^{h}_{as,i}$ indicates that the filter is less important and can be removed without significantly affecting the attention scores.

\subsection{Inter-head redundancy removal}
\label{sec:value_layer}

In the preceding section, we outlined the approach to preserving attention scores even with reduced embedding dimensions in the query and key layers.
Here, we introduce a pruning method for the value and other layers, such as FFN or patch embedding layer.

Previous works~\cite{attention_redundancy, michel2019sixteen} have revealed that a significant proportion of attention heads can be removed without causing significant performance deterioration. To remove this inter-head redundancy through neuron-level pruning, we propose to measure the distance between all the value layers of MSA module, irrespective of the heads. The importance score $\omega_{v,i}^{h}$ of the $i$-th filter in the value layer of the $h$-th head is as follows:
\begin{equation}
    \omega^{h}_{v,i} = \sum^{H}_{j=1}{\sum^{d_{v}}_{l=1}{(1-|\mathrm{cos}(W^{h}_{i}, W^{j}_{l})|)}}
    \label{eq:other_layers}
\end{equation}

The importance score of the value layer, denoted as $\omega_{v,i}^{h}$, indicates the redundancy of filter $i$ in the value layer compared to all value layers of the heads within the corresponding MSA layer. Consequently, filters with the lowest importance scores will be pruned in the initial stages of the pruning process.

\subsection{Accelerating Transformers}
\label{accelerating_transformer_based_model}

As described in Sec.~\ref{sec:preserving_attention_scores}, SNP removes the least correlated filter pairs ($Q_{i}^{h}, K_{i}^{h}$) to preserve attention scores and enhance the efficiency of the MSA module across various devices.

Most MSA modules, including MSA modules in DeiTs, are designed to compute all heads in parallel. 
To facilitate this parallel computation, SNP maintains consistency in the number of filter dimensions across heads, even as it independently selects filter indices for each head.

The green highlighted boxes in \cref{fig:proposed_methods} (b) represent layers connected by a single residual connection at the last add layer of the Transformer block. Unlike the matrix multiplication operator, the output of the residual connection becomes zero when all the interconnected layers return zero for specific filter indices. 
For this reason, the residual connection with masking always exceeds the performance of actual pruning, which restricts the set of possible pruning patterns~\cite{wan2023upscale}.

To accelerate the residual connection layers, all interconnected layers should be pruned identically. To achieve this, we sum up all the calculated importance scores for interconnected layers based on their filter indices, as shown in \cref{fig:proposed_methods} (e). Subsequently, we prune all the connected layers' least important filter indices.

\newcolumntype{C}{>{\centering\arraybackslash}X}
\begin{table*}[ht]
\caption{\textbf{Performance comparison of various pruning methods on ImageNet-1K.} The “Pruning" column represents the pruning methods that corresponding methods use to compress the MSA module. Each of the methods contains one of follows: head pruning (HP), block pruning (BP), neuron-level pruning (NP), neuron-level sparsity (NS).}
\centering
\begin{tabularx}{\textwidth}{C|C|C|CCCC}
\hline \hline
                                                        & \textbf{Method}                       & \textbf{Pruning}  & \textbf{Top-1 (\%)}  & \textbf{Top-5 (\%)}  & \textbf{GFLOPs}    & \textbf{Params (M)} \\ \hline
\multicolumn{1}{c|}{\multirow{6}{*}{\textbf{DeiT-Tiny}}}   & Original~\cite{touvron2021training_deit}& -               & 72.20            & 91.10           & 1.3                & 5.7                 \\
\multicolumn{1}{c|}{}                                   & SSViTE~\cite{ssvite}                  & HP                & 70.12           & -               & 0.9                & 4.2                 \\
\multicolumn{1}{c|}{}                                   & WDPruning~\cite{WDPruning}            & HP+BP            & 70.34            & 89.82           & 0.7                & 3.5                 \\
\multicolumn{1}{c|}{}                                   & X-Pruner~\cite{xpruner}               & HP                & 71.10            & 90.11           & 0.6                & -                   \\
\multicolumn{1}{c|}{}                                   & \textbf{SNP}                          & \textbf{NP}       & \textbf{70.29}  & \textbf{90.01}  & \textbf{0.6}       & \textbf{3.0}       \\
\multicolumn{1}{c|}{}                                   & UVC~\cite{uvc}                        & HP+NS+BP        & 70.60            & -               & 0.5                & -                   \\ \hline
\multicolumn{1}{c|}{\multirow{7}{*}{\textbf{DeiT-Small}}}  & Original~\cite{touvron2021training_deit}& -                & 79.85           & 95.00           & 4.6                & 22.1                \\
\multicolumn{1}{c|}{}                                   & SSViTE~\cite{ssvite}                  & HP                & 79.22           & -               & 3.1                & 14.6                \\
\multicolumn{1}{c|}{}                                   & WDPruning~\cite{WDPruning}            & HP+BP            & 78.38           & 94.05           & 2.6                & 13.3                \\
\multicolumn{1}{c|}{}                                   & X-Pruner~\cite{xpruner}                & HP               & 78.93           & 94.24           & 2.4                & -                   \\
\multicolumn{1}{c|}{}                                   & UVC~\cite{uvc}                        & HP+NS+BP        & 78.82           & -               & 2.3                & -                   \\
\multicolumn{1}{c|}{}                                   & \textbf{SNP}                          & \textbf{NP}       & \textbf{78.52}  & \textbf{94.37}  & \textbf{2.0}       & \textbf{10.0}      \\
\multicolumn{1}{c|}{}                                   & \textbf{SNP}                          & \textbf{NP}       & \textbf{73.32}  & \textbf{91.66} & \textbf{1.3}       & \textbf{6.4}        \\ \hline
\multicolumn{1}{c|}{\multirow{6}{*}{\textbf{DeiT-Base}}}   & Original~\cite{touvron2021training_deit}   & -            & 81.80           & 95.59           & 17.6               & 86.6                \\
\multicolumn{1}{c|}{}                                   & SSViTE~\cite{ssvite}                  & HP                & 82.22           & -               & 11.8               & 56.8                \\
\multicolumn{1}{c|}{}                                   & WDPruning~\cite{WDPruning}            & HP+BP            & 80.76           & 95.36           & 9.9                & 55.3                \\
\multicolumn{1}{c|}{}                                   & X-Pruner~\cite{xpruner}                & HP               & 81.02           & 95.38           & 8.5                & -                   \\
\multicolumn{1}{c|}{}                                   & UVC~\cite{uvc}                        & HP+NS+BP        & 80.57           & -               & 8.0                  & -                   \\
\multicolumn{1}{c|}{}                                   & \textbf{SNP}                          & \textbf{NP}       & \textbf{79.63}  & \textbf{94.37}  & \textbf{6.4}      & \textbf{31.6}      \\
\hline \hline
\end{tabularx}
\label{table:main_table}
\end{table*}



\section{Experiments}

To ensure a fair comparison with existing methods, we apply SNP to prune DeiT~\cite{touvron2021training_deit} architectures trained on the ImageNet-1K~\cite{deng2009imagenet} dataset. 
Additionally, we conduct experiments on the efficient Transformer model, EfficientFormer-L1, to confirm the robustness of the SNP. Furthermore, a series of ablation studies are conducted to gain a comprehensive understanding of our methodology.

\subsection{Implementation details}

The overall pruning and fine-tuning process are executed on the pre-trained DeiT\footnote{https://github.com/facebookresearch/deit} and EfficientFormer-L1\footnote{https://github.com/snap-research/EfficientFormer} released from the official implementation on ImageNet-1K.
Throughout the fine-tuning phase of the pruned model, we maintain consistent settings across all models, except for the batch size and learning rate. The batch size is set to 256, and to prevent weight explosion, we adjust the learning rate of the compressed model to 1/10 or 1/100 of the original model.

To evaluate the reduced latency using SNP, we have configured four testing scenarios: one on a CPU and another on GPU for both edge devices and server processors. 
We employ a standard PyTorch model for profiling on the server processors (Intel Xeon Silver 4210R and NVIDIA GeForce RTX 3090). Profiling on the Raspberry Pi 4B and Jetson Nano is conducted using the ONNX and TensorRT formats, respectively.
All latencies are measured using a single image as an input, except for the GPU of the server processor (RTX 3090), where it is set to 64 images.

\begin{table*}[ht]
\caption{\textbf{Inference speed and Top-1 accuracy of the compressed model across different devices.} Performance evaluation involves accuracy on ImageNet-1K and inference time for the compressed DeiTs and EfficientFormer-L1. Latency is benchmarked with 200 warm-up runs and averaged over 1000 runs. In latency measurement, a single image is used as the batch size, except for the RTX 3090, where 64 images are employed in a single batch.}
\centering
\resizebox{\textwidth}{!}{
\begin{tabular}{ccccccc}
\hline \hline
\multirow{2}{*}{Model}& \multirow{2}{*}{Top-1 (\%)} & \multirow{2}{*}{GFLOPs} & \multicolumn{2}{c}{Edge devices (ms)}    & \multicolumn{2}{c}{Server processors (ms)}  \\  \cmidrule(lr){4-5} \cmidrule(lr){6-7}
                    &                       &                   & Raspberry Pi 4B (.onnx)           & Jetson Nano (.trt)            & Xeon Silver 4210R (.pt)     & RTX 3090 (.pt)             \\ \hline
DeiT-Tiny              & 72.20                 & 1.3               & 139.13                           & 41.03                         & 34.74                         & 18.65                          \\
\textbf{+ SNP (Ours)}        & \textbf{70.29}        & \textbf{0.6}      & \textbf{81.63 (1.70$\times$)}    & \textbf{26.67 (1.54$\times$)} & \textbf{25.25 (1.38$\times$)} & \textbf{17.82 (1.05$\times$)}  \\ \hline
DeiT-Small              & 79.80                 & 4.6               & 401.27                           & 99.32                         & 53.37                         & 46.13                          \\
\textbf{+ SNP (Ours)}        & \textbf{78.52}        & \textbf{2.0}      & \textbf{199.15 (2.01$\times$)}   & \textbf{45.51 (2.18$\times$)} & \textbf{38.57 (1.38$\times$)} & \textbf{32.91 (1.40$\times$)}  \\
\textbf{+ SNP (Ours)}        & \textbf{73.32}        & \textbf{1.3}      & \textbf{136.68 (2.94$\times$)}   & \textbf{32.03 (3.10$\times$)} & \textbf{33.46 (1.60$\times$)} & \textbf{26.98 (1.71$\times$)}  \\ \hline
DeiT-Base              & 81.80                 & 17.6              & 1377.71                          & 293.29                        & 122.03                        & 151.35                         \\
\textbf{+ SNP (Ours)}        & \textbf{79.63}        & \textbf{6.4}      & \textbf{565.68  (2.44$\times$)}  & \textbf{132.55 (2.21$\times$)}& \textbf{64.65 (1.89$\times$)} & \textbf{72.96 (2.07$\times$)}  \\ 
\textbf{+ SNP (Ours) + Head}        & \textbf{79.12}        & \textbf{3.5}      & \textbf{307.00  (4.48$\times$)}  & \textbf{59.47 (4.93$\times$)}& \textbf{46.09 (2.65$\times$)} & \textbf{39.31 (3.85$\times$)}  \\ \hline
EfficientFormer-L1  & 79.20                 & 1.3               & 169.13                           & 30.95                         & 43.75                         & 26.19                          \\
\textbf{+ SNP (Ours)}        & \textbf{75.53}        & \textbf{0.6}      & \textbf{95.12 (1.78$\times$)}    & \textbf{19.78 (1.56$\times$)} & \textbf{38.25 (1.14$\times$)} & \textbf{17.24 (1.52$\times$)}  \\
\textbf{+ SNP (Ours)}        & \textbf{74.51}        & \textbf{0.5}      & \textbf{82.60 (2.05$\times$)}    & \textbf{17.76 (1.74$\times$)} & \textbf{35.15 (1.24$\times$)} & \textbf{16.01 (1.64$\times$)}  \\ 
\hline \hline
\end{tabular}}
\label{table:latency}
\end{table*}



\subsection{Quantitative results}

\subsubsection{Comparison with other methods}

Despite the constraints outlined in Sec.~\ref{accelerating_transformer_based_model}, SNP not only maintains accuracy comparable to existing methods but also significantly reduces inference time across diverse hardware and data types on the ImageNet-1K dataset, as shown in \cref{table:main_table} and \cref{table:latency}.

In a recent study~\cite{wan2023upscale}, unconstrained masking generally outperforms post-training accuracy of pruned models by an average of 2.1\% on ImageNet-1K. 
Compared to unconstrained head masking approaches like SSVITE~\cite{ssvite} and X-Pruner~\cite{xpruner}, SNP achieves significantly higher compression rates for all DeiTs FLOPs (30.64\% and 10.64\%) with minimal performance degradation (0.83\% and 0.87\%), much less than the 2.1\% average mentioned.

Furthermore, compared to other pruning approaches like WDPruning~\cite{WDPruning} and UVC~\cite{uvc}, which use a combination of pruning techniques to compress DeiTs, SNP achieves comparable accuracy solely through neuron-level pruning. Notably, DeiT-Tiny with SNP outperforms WDPruning by 0.14\%, with the removal of 3.3 million parameters and a reduction of 0.6 GFLOPs. Compared to UVC, SNP exhibits negligible performance degradation, averaging 0.51\% across all DeiTs while using 7.65\% fewer FLOPs.

\subsubsection{Large compressed vs. Small hand-crafted}

DeiT-Small with SNP, a large pruned model, outperforms the smaller, hand-crafted DeiT-Tiny in both accuracy and latency, achieving a notable 1.12\% improvement in top-1 accuracy while maintaining similar FLOPs. 
Additionally, DeiT-Small with SNP exhibits enhanced speed compared to the original DeiT-Tiny across edge devices and CPU-based server processors.
Notably, its speed increases up to 21.94\% compared to the original DeiT-Tiny running on Jetson Nano, a GPU-based edge device.
This substantial performance gap underscores the superiority of the compressed model (DeiT-Small with SNP) over the smaller hand-crafted designed model (DeiT-Tiny) in both overall performance and speed.
\begin{figure}[t]
    \centering
    \includegraphics[width=\columnwidth]{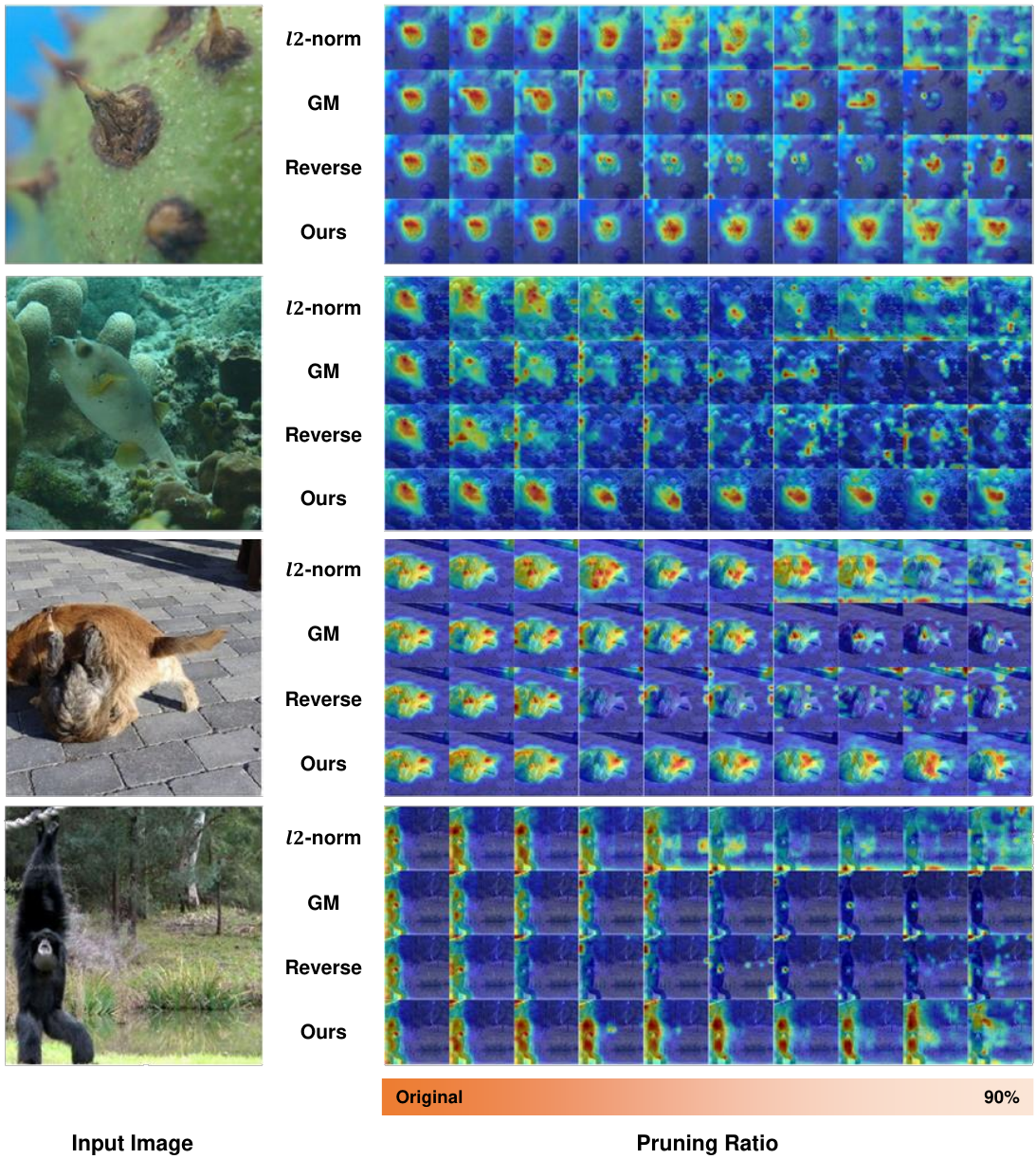}
    \caption{\textbf{Attention maps with varying pruning criteria and compression ratios.}~All query and key layers are locally pruned based on the specified pruning ratio without fine-tuning. The importance scores of $l2$-norm and GM on query and key layers are combined by filter index and pruned simultaneously. “Reverse" represents the reverse order of SNP.}
    \label{fig:atentionmap_by_criteria_and_pruning_ratio}
\end{figure}

\begin{figure*}[ht]
    \centering
    \includegraphics[scale=0.8]{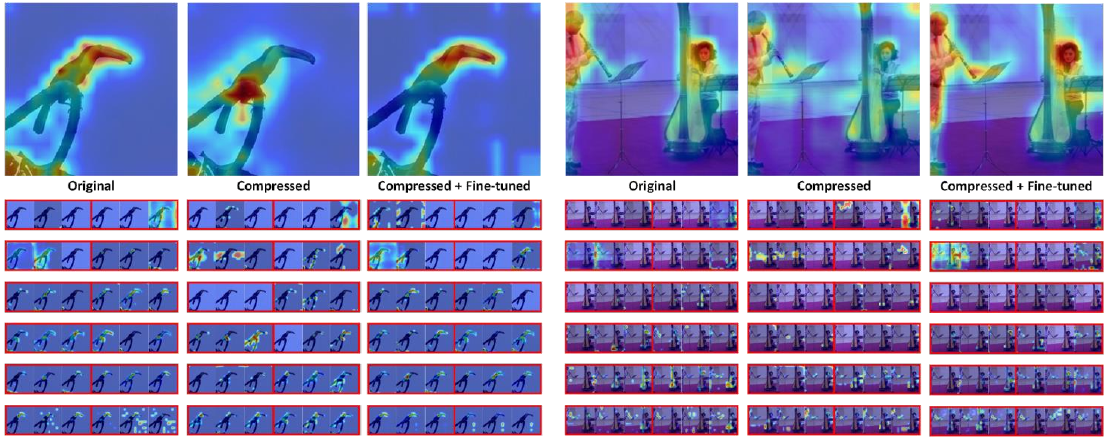}
    \caption{\textbf{Attention maps from the original, compressed, and fine-tuned DeiT-Tiny with SNP.} The attention maps in the first row are visualized using the attention rollout~\cite{attention_rollout}. Each red box contains three attention maps from each head of the MSA module, ordered accordingly.}
    \label{fig:all_attention_map}
\end{figure*}

\subsubsection{Accelerating Transformer-based models}
As depicted in ~\cref{table:latency}, SNP achieves impressive acceleration of DeiTs by a factor of 1.44$\times$ to 2.44$\times$ on edge devices and 1.05$\times$ to 2.07$\times$ on server processors. 
This acceleration is notable, with negligible average performance degradation of 1.79\%, specifically 1.91\%, 1.28\%, and 2.17\% for DeiT-Tiny, DeiT-Small, and DeiT-Base, respectively.

Compared to WDPruning~\cite{WDPruning}, which employs both head and block pruning, SNP surpasses in terms of latency for all of DeiTs on CPUs and GPUs except for the smallest DeiT model DeiT-T as ~\cref{table:latency_comparisoin}, even though SNP employs neuron-level pruning only. 
SNP accelerates the original DeiTs by 1.38$\times$ and 2.07$\times$, while WDPruning achieves a comparatively modest acceleration of 1.18$\times$.

Since SNP reduces the number of filters instead of removing operators such as matrix multiplication, its superiority becomes more evident when linear or convolutional layers constitute a larger proportion of the original model's computation time, particularly on GPU, where parallel computing is feasible. SNP accelerates the original DeiTs by 1.05$\times$ to 2.07$\times$ as the size of the DeiT model grows. 
In contrast, WDPruning, which involves the removal of entire layers and associated operators of both head and block, shows modest acceleration of 1.18$\times$ on all of DeiT models.

To ascertain the robustness of SNP across various Transformer models, especially on the efficiently designed model, we conducted additional experiments on EfficientFormer-L1. SNP accelerates the model by 1.78$\times$ and 2.05$\times$ faster on Raspberry Pi 4B and 1.56$\times$ and 1.74$\times$ faster on Jetson Nano. In particular, the compressed EfficientFormer-L1 achieves an acceptable accuracy of 75.53\% and 74.51\%.

\begin{table}[ht]
\caption{\textbf{Inference speed comparison with the conventional head and block pruning approach, WDPruning.} 
As the device verifying the efficiency of the proposed method are different, we performed the performance comparison based on how much each method accelerates the original model.}
\centering
\resizebox{\columnwidth}{!}{
\begin{tabular}{ccccc}
\hline \hline
                    & Top-1 (\%)            & GFLOPs        & CPU                    & GPU  \\ \hline
DeiT-T              & 72.20                 & 1.3           & 1.00$\times$          & 1.00$\times$ \\
WDPruning~\cite{WDPruning}           & 70.34                 & 0.7           & 1.25$\times$          & 1.19$\times$ \\
\textbf{SNP}        & \textbf{70.29}        & \textbf{0.6}  & \textbf{1.38$\times$} & \textbf{1.05$\times$} \\ \hline
DeiT-S              & 79.80                 & 4.6           & 1.00$\times$          & 1.00$\times$ \\
WDPruning~\cite{WDPruning}           & 78.38                 & 2.6           & 1.21$\times$          & 1.18$\times$ \\
\textbf{SNP}        & \textbf{78.52}        & \textbf{2.0}  & \textbf{1.38$\times$} & \textbf{1.40$\times$} \\ \hline
DeiT-B              & 81.80                 & 17.6          & 1.00$\times$          & 1.00$\times$ \\
WDPruning~\cite{WDPruning}           & 80.76                 & 9.9          & 1.32$\times$          & 1.18$\times$ \\
\textbf{SNP}        & \textbf{79.63}        & \textbf{6.4}  & \textbf{1.89$\times$} & \textbf{2.07$\times$} \\
\hline \hline
\end{tabular}}
\label{table:latency_comparisoin}
\end{table}

\newcolumntype{Y}{>{\centering\arraybackslash}X}
\newcolumntype{Z}{>{\centering\arraybackslash\hsize=.2\textwidth}X}
\begin{table*}[ht!]
\caption{\textbf{Performance of SNP without fine-tuning across different data quantities at various pruning ratios.} To determine the proper number of images for calculating the importance score, we conduct local pruning on the query and key layers at pruning ratios of 10\%, 30\%, 50\%, 70\%, and 90\%. All performance metrics are assessed without fine-tuning. The latency is measured on Raspberry Pi 4B.}
\centering
\begin{tabularx}{0.9\textwidth}{ZYYYYYY}
\hline \hline
\multirow{2}{*}{\textbf{Number of images}} & \multicolumn{6}{c}{\textbf{Performance by pruning ratio}} \\ \cmidrule(lr){2-7}
                &\textbf{Original}  & \textbf{10\%}         & \textbf{30\%}         & \textbf{50\%}         & \textbf{70\%}         & \textbf{90\%}       \\ \hline 
\textbf{1}      &72.20               & 71.15                 & 67.96                 & 57.60                 & 38.68                 & 11.17    \\ 
\textbf{4}  &72.20               & 70.82                 & 67.86                 & 57.65                 & 38.05                & 11.86    \\
\textbf{16} &72.20               & 70.72                 & 67.42                 & 58.60                 & 39.46                & 9.11      \\
\textbf{64} &72.20               & 70.83                 & 67.55                 & 59.50                 & 42.13                & 12.83     \\
\textbf{256} &72.20               & 70.75                & 68.14                 & 59.58                 & 42.70                & 14.48     \\
\hline 
\textbf{Latency (ms)}   &139.13  & 133.60                & 129.29                & 119.41                & 117.48                & 112.32     \\
\textbf{Params (M)}     &5.7    & 5.59                  & 5.41                  & 5.23                  & 5.06                  & 4.88     \\ \hline \hline
\end{tabularx}
\label{table:num_of_imgs}
\end{table*}


\subsection{Qualitative results}

In the subsequent sections, attention rollout~\cite{attention_rollout} is employed to randomly selected images from the test datasets to visualize the attention maps of the DeiTs and assess the efficacy of SNP. This evaluation unfolds across two key facets:
\begin{itemize}
    \item \textbf{Preserving the attention scores}: Visualize attention maps to verify the effectiveness of the proposed criteria in preserving attention scores.
    \item \textbf{Restoring the attention scores after SNP}: Visualize attention maps to find the effectiveness of SNP in restoring attention scores.
\end{itemize}


\subsubsection{Preserving the attention scores}
\label{sec:viz_attention_scores}

To visualize SNP effectiveness, we apply four pruning criteria to compress the query and key filter pairs in all MSA modules of DeiT-Tiny: SNP, $l2$-norm, geometric median (GM)~\cite{gm_mask_0}, and “Reverse".

The “Reverse" criterion prioritizes pruning the most important filter pairs first, keeping the least important pairs until the end. However, both $l2$-norm and GM pruning criteria ignore the graphical connectivity of the MSA module, independently evaluating importance scores for query and key layers. To handle identical filter indices during pruning, we aggregate scores based on filter indices, removing the least important indices from both layers.

In \cref{fig:atentionmap_by_criteria_and_pruning_ratio}, attention maps for the original DeiT-Tiny and locally pruned models are presented across various pruning ratios (10\% to 90\% with 10\% intervals). 
Our proposed method effectively maintains the original attention map even after pruning over 80\% of the filters, whereas other methods show fragmented attention maps at much lower pruning ratios (typically 30\% or less). 
These results highlight the potential of our neuron-level pruning criteria, utilizing SVD to preserve attention scores, in reducing the size and speeding up the execution of MSA modules without compromising accuracy.

\subsubsection{Restoring the attention scores after SNP}

\cref{fig:all_attention_map} illustrates the overall and per-head attention maps of the original, compressed, and fine-tuned DeiT-Tiny, respectively.
The first row shows the overall attention maps of the respective models. Notably, the compressed model maintains a well-preserved overall attention map, despite pruning all layers, including values and FFN, resulting in a 53\% reduction in FLOPs and a 46\% reduction in parameters. 
Especially, we can observe that the attention map is well-restored after the fine-tuning process.

The twelve red boxes below the overall attention map depict per-head attention maps for twelve layers of each original, compressed, and fine-tuned models respectively. As shown in \cref{fig:all_attention_map}, it is evident that the attention maps for each head are effectively preserved and restored in each of the compressed and fine-tuned models.

\subsection{Ablation Studies}
\subsubsection{Importance scores by the data quantity}

Since attention scores are influenced by input, as demonstrated in Eq.~(\ref{eq:attention_score}), the suggested importance scores for query and key filter pairs  (\cref{eq:cos_sim}) susceptible to the distribution of the input image $X$ 
To validate the method's robustness, we compute importance scores using various image quantities, pruning query and key layers at different ratios, without fine-tuning process.

As depicted in \cref{table:num_of_imgs}, the SNP demonstrates a slight advantage in preserving performance with an increasing number of images, outperforming models compressed with fewer images as the compression ratio increases. 
However, this improvement comes at the cost of increased computational time for SNP calculations. 
Considering these factors into account, we opt to use 64 images, which yield the second-best performance among the given number of images. This decision strikes a balance between achieving satisfactory performance and maintaining computational efficiency.

\subsubsection{Performance comparison across pruning ratios}

To assess the robustness of SNP, we examine the performance of DeiT-Tiny under various pruning criteria and different pruning ratios, as described in ~\cref{sec:viz_attention_scores}.

\begin{figure}[t]
    \centering
    \includegraphics[width=0.90\columnwidth]{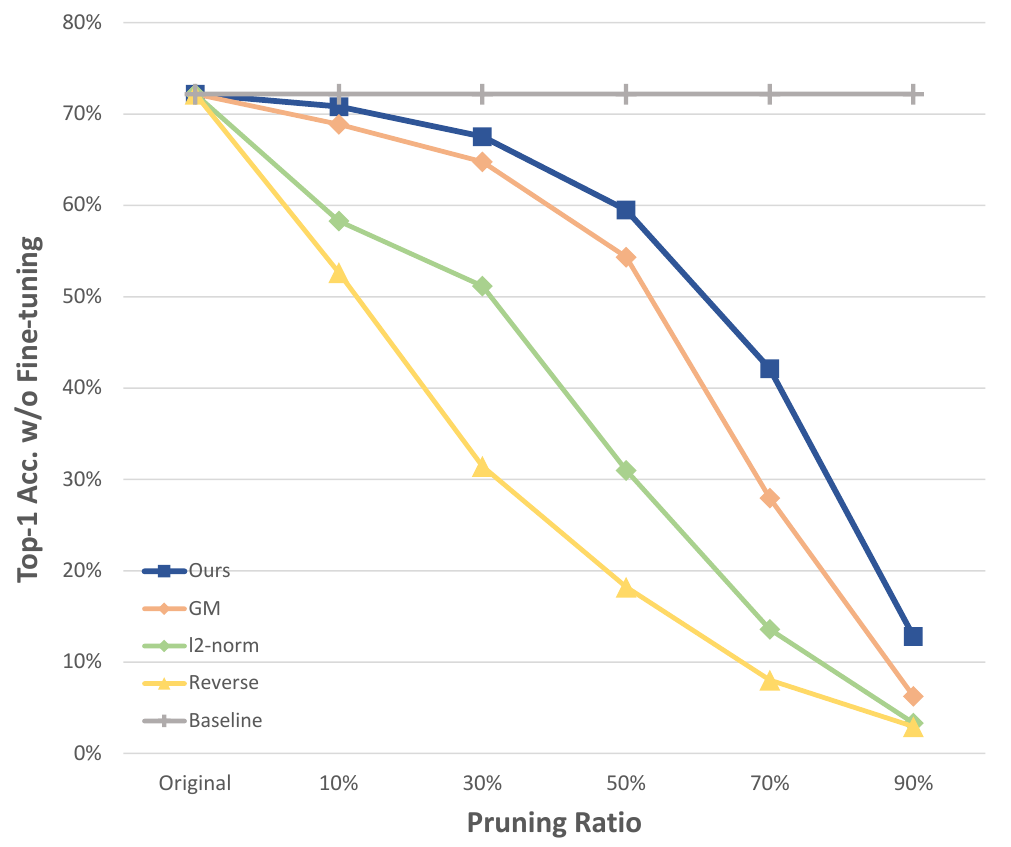}
    \caption{\textbf{Top-1 accuracy of compressed DeiT-Tiny on ImageNet using several pruning criteria without fine-tuning.} Query and key layers are locally pruned using various pruning criteria : SNP, GM, $l2$-norm, reverse order of SNP (“Reverse"), and original DeiT-Tiny (“Baseline").}
    \label{fig:local_perf_wo_train}
\end{figure}

As illustrated in \cref{fig:local_perf_wo_train}, SNP consistently outperforms other pruning criteria across all pruning ratios. In contrast, models compressed with “Reverse" criteria exhibit the lowest performance at all pruning ratios, underscoring the robustness of the proposed approach.

The figures \cref{fig:atentionmap_by_criteria_and_pruning_ratio} and \cref{fig:local_perf_wo_train} both validate the effectiveness of SNP in maintaining attention scores from both numerical and qualitative perspectives.
Despite a pruning ratio of 80\% and the absence of fine-tuning, SNP is able to keep the original attention scores intact and surpasses other pruning criteria in performance.

\subsubsection{SNP with other pruning approaches}
Since, SNP is the first neuron-level pruning approach to accelerate not only MSA modules but also individual heads by eliminating graphically linked filter indices, it paves the way for potentially enhancing the speed of ViT models in the future by integrating SNP with other pruning techniques.

To demonstrate this, we conduct additional experiments combining head pruning with SNP as depicted in ~\cref{table:latency}. We utilize Equation \ref{eq:other_layers} to measure the importance score of each head, and then 50\% of the heads were selected for removal. 
Using SNP with head pruning, it can reduce 80\% of the parameters and computational costs, but with a 2.68\% performance degradation.
The corresponding compressed model achieves 307ms on Raspberry Pi 4B, 59.47ms on Jetson Nano, 46.09ms on Xeon Silver 4201R, and 39.31ms for RTX3090, which is 4.48$\times$, 4.93$\times$, 2.65$\times$, and 3.85$\times$ faster than the original model, respectively.

Yet, combining traditional neuron-level sparsity methods (called “masking") with head pruning fails to yield additional acceleration beyond what is attainable through head pruning alone. 
Thus, with this paper, we hope to explore various pruning techniques in combination to obtain more efficient models for both performance and speed in the future.

\section{Conclusion}

In this paper, we propose a novel graph-aware neuron-level pruning method, SNP, designed to compress and accelerate Transformer-based models. SNP proposes two pruning criteria for preserving attention scores and eliminating inter-head redundancy.
Using SNP, a large compressed model outperforms small, hand-crafted designed models in both performance and latency on edge devices. Moreover, the compressed models exhibit astonishing results in latency on various devices, with negligible performance degradation.

As this work is a first attempt to accelerate MSA modules using neuron-level pruning alone, many challenges remain. One is to incorporate other pruning methods, such as head or block pruning for a more efficient Transformer model. Another challenge is to apply SNP to other vision tasks, including image generation, which requires high computational costs on both training and inference.

We believe that these works encourage the adoption of model pruning as a tool to improve both the applicability of ViTs in resource-constrained environments and to reduce the training costs of large models by integrating with the training process.

{
    \small
    \bibliographystyle{ieeenat_fullname}
    \bibliography{main}
}

\clearpage
\setcounter{page}{1}
\maketitlesupplementary

\setcounter{section}{0}
\renewcommand{\thesection}{\Alph{section}}
\setcounter{equation}{0}
\renewcommand{\theequation}{\Alph{equation}}
\setcounter{table}{0}
\renewcommand{\thetable}{\Alph{table}}
\setcounter{figure}{0}
\renewcommand{\thefigure}{\Alph{figure}}

\section{Accelerating Transformer models}
\label{sup:Accelerating_trasnformer}

\subsection{Accelerating MSA modules}

The implementation of MSA modules varies significantly among developers. 
Some approaches involve implementing each head's query, key, and value layers independently. 
In contrast, other implementations consolidate all of the head's query, key, and value layers into a single layer, utilizing the reshape operator to accelerate the MSA module through parallel computation.

The MSA module of DeiTs\footnote{https://github.com/facebookresearch/deit} is implemented as described in ~\cref{fig:msa_implementation} (a). 
All linear layers within the MSA module are consolidated into a single linear layer with a shape of $\mathbb{R}^{d\times H*(d_{q}+d_{k}+d_{v})}$. 
The output matrix of the consolidated linear layer, with a shape of $\mathbb{R}^{N\times H*(d_{q}+d_{k}+d_{v})}$, is reshaped into $\mathbb{R}^{N\times 3 \times H \times d_{q}}$, where 3 represents three layers: query, key, and value. 
This reshaped output matrix is subsequently divided into individual sets corresponding to each role of linear layers (query, key, and value). 
The conventional self-attention process is then conducted using the output of each role in a parallel manner.

Due to the reshape operator in the MSA implementation mentioned above, neuron-level pruning encounters two constraints to accelerate the MSA module:
\begin{itemize}
    \item \textbf{Constraints on Q, K, and V:} All query, key, and value layers of self-attention module  must have the same dimensions.
        \begin{itemize}
            \item $d_{q} = d_{k} = d_{v}$
        \end{itemize}
    \item \textbf{Constraints on multi-head:} All heads in the MSA module should have the same dimension.
        \begin{itemize}
            \item $d_{\{q,k,v\}}^{1} = d_{\{q,k,v\}}^{2} = ... = d_{\{q,k,v\}}^{H}$
        \end{itemize}
\end{itemize}

\subsubsection{Constraints on Q, K, and V}

The original implementation of the MSA module in DeiTs enforces an equal number of dimensions for all query, key, and value layers. 
However, SNP proposes to remove identical filter indices for the graphically connected query and key layers while independently removing the value layer. 
This necessitates overcoming constraints on Q, K, and V.

\begin{figure*}[t]
    \centering
    \begin{subfigure}[b]{\columnwidth}
        \centering
        \includegraphics[scale=0.35]{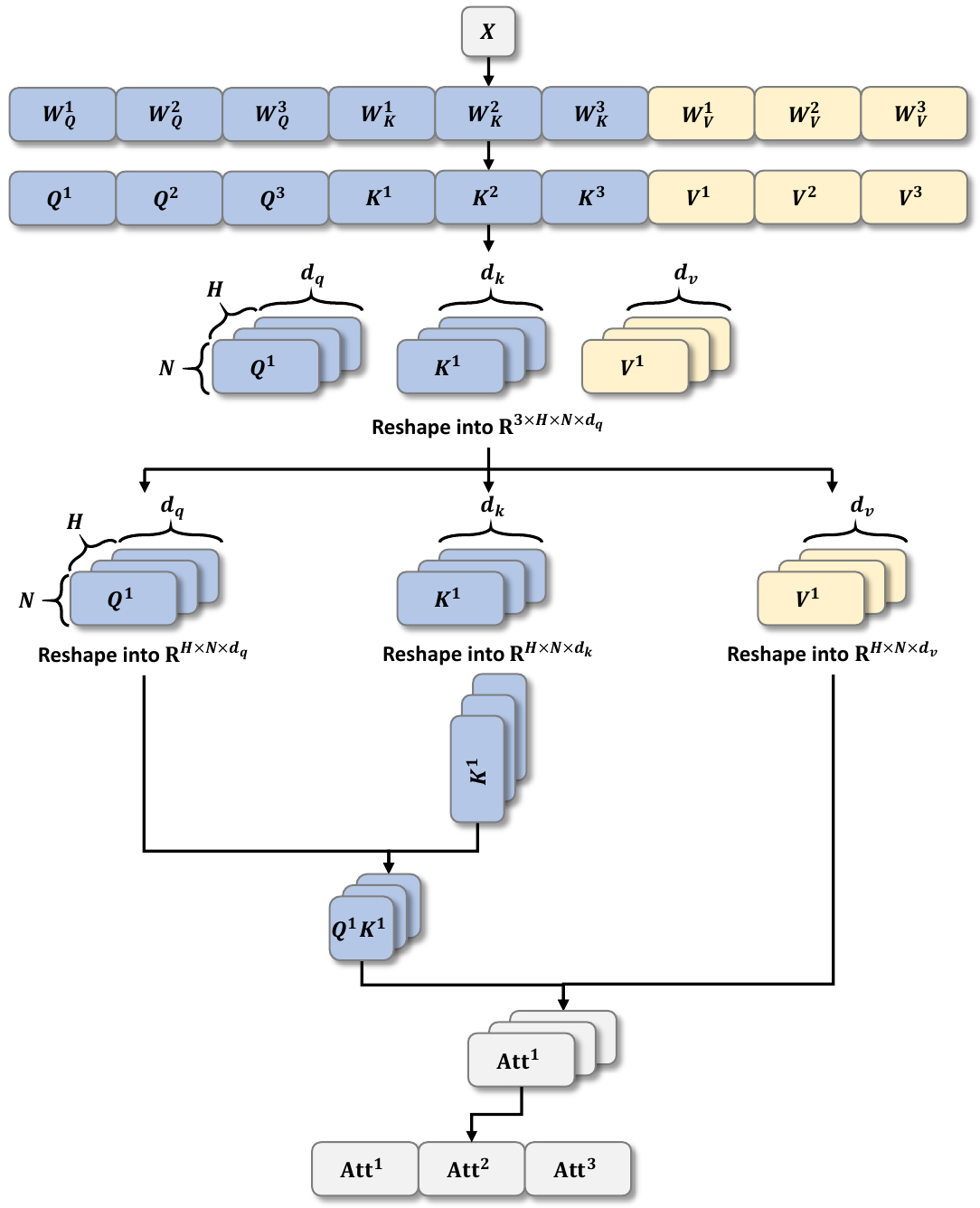}
        \caption{\textbf{Detailed implementation of MSA module in DeiTs.}}
        \label{fig:sup_msa_timm}
    \end{subfigure}
    \hfill
    \begin{subfigure}[b]{\columnwidth}
        \centering
        \includegraphics[scale=0.35]{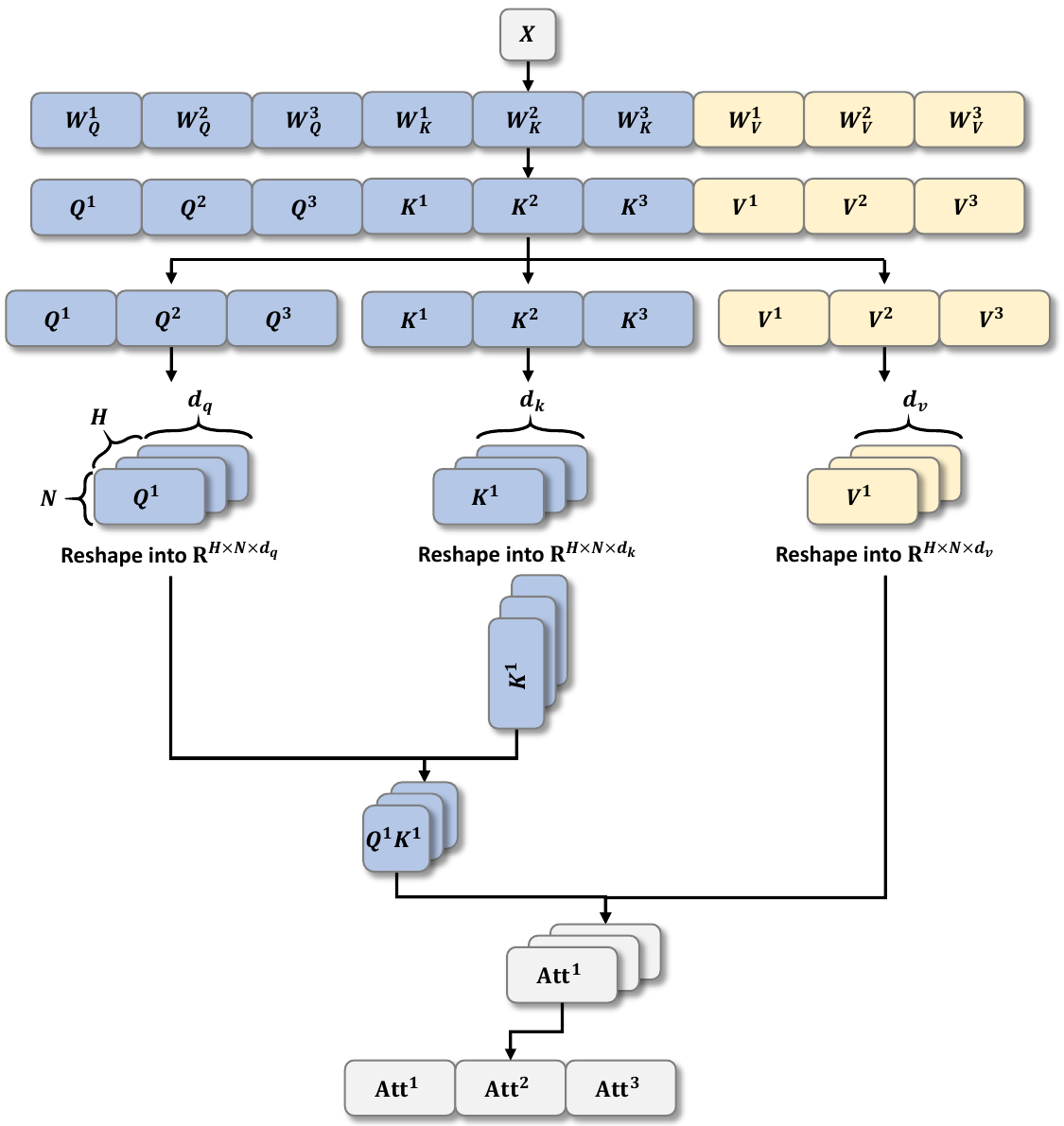}
        \caption{\textbf{Detailed implementation of MSA module in SNP.}}
        \label{fig:sup_msa_ours}
    \end{subfigure}
    \caption{\textbf{Detailed implementation of MSA module on original DeiTs and SNP.}
        \textbf{(a) MSA module of the original DeiTs.}  
        Reshape operator transforms the output matrix into $3 \times H \times N \times d_{q}$, followed by a split layer to divide the first dimension, where $3$ represents query, key, and value layers.
       \textbf{(b) Converted MSA module in SNP.} 
       Split operator follows the linear layer to ensure a distinction between the query, key, and the value dimension. 
       For parallel computation by head, the reshape operator comes after the split layer and follows the ordinary MSA module in sequence.
    }
    \label{fig:msa_implementation}
\end{figure*}

To address this, a reshape layer followed by a split layer is converted into the split layer and reshape layer, as shown in ~\cref{fig:msa_implementation} (b). 
Specifically, the split layer divides the output matrix from a single linear layer by its role (query, key, and value), and the reshape layer transforms each matrix $\mathbb{R}^{N\times (H*d_{\{q,k,v\}})}$ into a matrix shape of $\mathbb{R}^{N\times H \times d_{\{q,k,v\}}}$. 
Each reshaped matrix follows the original MSA module's workflows, such as scaled dot matrix multiplication, etc.

Compared to the original implementation of the MSA module in DeiTs, MSA module with SNP includes two additional reshape layers, which may impact the model's latency. 
To assess this impact, we replaced all MSA modules in DeiTs with the MSA module of SNP and measured the inference time.
As shown in \cref{tab:converted_msa}, there is no significant difference between the original model and the converted model. 
The converted DeiT-Tiny is slower than the original model by 0.72 ms, and the DeiT-Base model is slower by 0.43 ms, while the converted DeiT-Small is slightly faster than the original model. 
By changing the order of the split layer and the reshape layer, SNP overcomes the first constraints, albeit with a minor difference in inference time compared to the original model.

\begin{table}[t]
\caption{\textbf{Inference time of original DeiTs vs. converted DeiTs.} Profiling of the original and converted DeiTs is conducted using standard PyTorch file types (.pt) on RTX 3090. The averaged inference time is measured based on 1,000 runs after 200 warm-up runs.
A single batch with 64 images is utilized for the evaluation.}
\centering
\resizebox{\columnwidth}{!}{
\begin{tabular}{cccc}
\hline\hline
                 & \multicolumn{3}{c}{Latency (ms)} \\ \cmidrule(lr){2-4}
                 & DeiT-Tiny    & DeiT-Small    & DeiT-Base \\ \hline
Original DeiTs   & 18.65        & 46.13         & 151.35 \\
DeiTs with SNP   & 19.37        & 46.12         & 151.78 \\
\hline\hline
\end{tabular}}
\label{tab:converted_msa}
\end{table}

\begin{table*}[htbp]
\caption{\textbf{Pruning indices of the first MSA module of compressed DeiT-Tiny using SNP.} Graphically connected query and key pairs “matmul" are pruned by $40\%$ identically, while value layer “matmul$\_1$" is pruned $20\%$ independently.}
\centering
\resizebox{\textwidth}{!}{
\begin{tabular}{cc|ccccccccccccccccccccccccc}
\hline \hline
        &       &\multicolumn{25}{c}{\textbf{Order of pruning indices}}\\ 
        &       & 1  & 2  & 3  & 4  & 5  & 6  & 7  & 8  & 9  & 10 & 11 & 12 & 13 & 14 & 15 & 16 & 17 & 18 & 19 & 20 & 21 & 22  & 23& 24 & 25  \\ \hline
Head~\#1 & Query & 45 & 24 & 1  & 18 & 2  & 3  & 32 & 9  & 58 & 46 & 25 & 19 & 41 & 17 & 22 & 39 & 16 & 53 & 50 & 59 & 36 & 0  & 51 & 12 & 5  \\
        & Key   & 45 & 24 & 1  & 18 & 2  & 3  & 32 & 9  & 58 & 46 & 25 & 19 & 41 & 17 & 22 & 39 & 16 & 53 & 50 & 59 & 36 & 0  & 51 & 12 & 5  \\
        & Value & 2  & 8  & 21 & 54 & 26 & 55 & 59 & 61 & 33 & 47 & 6  & 13 &  &    &    &    &    &    &    &    &    &    &    &    &    \\ \hline
Head~\#2 & Query & 51 & 42 & 18 & 27 & 44 & 28 & 25 & 3  & 21 & 12 & 61 & 62 & 10 & 53 & 15 & 34 & 5  & 23 & 40 & 2  & 54 & 46 & 6  & 0  & 56 \\
        & Key   & 51 & 42 & 18 & 27 & 44 & 28 & 25 & 3  & 21 & 12 & 61 & 62 & 10 & 53 & 15 & 34 & 5  & 23 & 40 & 2  & 54 & 46 & 6  & 0  & 56 \\
        & Value & 37 & 23 & 0  & 53 & 33 & 5  & 41 & 61 & 15 & 25 & 47 & 18 &  &    &    &    &    &    &    &    &    &    &    &    &    \\\hline
Head~\#3 & Query & 14 & 7  & 63 & 38 & 52 & 22 & 32 & 29 & 11 & 26 & 12 & 27 & 35 & 58 & 1  & 51 & 55 & 43 & 0  & 34 & 45 & 56 & 59 & 30 & 48 \\
        & Key   & 14 & 7  & 63 & 38 & 52 & 22 & 32 & 29 & 11 & 26 & 12 & 27 & 35 & 58 & 1  & 51 & 55 & 43 & 0  & 34 & 45 & 56 & 59 & 30 & 48 \\
        & Value & 54 & 1  & 11 & 31 & 35 & 13 & 36 & 45 & 26 & 23 & 5  & 44 &  &    &    &    &    &    &    &    &    &    &    &    &    \\ \hline \hline
\end{tabular}}
\label{tab:prune_indices}
\end{table*}

\subsubsection{Constraints on multi-head}

In the MSA module, each head contains various operators, such as matrix multiplication and softmax, which require relatively long computation times.
For this reason, most developers implement the MSA module in a parallel manner.
To preserve the parallel implementation of the original MSA module, SNP prunes an equal number of filters for each of the query, key, and value layers across all heads but with different filter indices, as shown in \cref{tab:prune_indices}. 
This approach illustrates that neuron-level pruning not only accelerates the single-head self-attention module but also enhances the efficiency of Transformer models on any devices.

\subsection{Accelerating residual connection}
\label{sec:sup_accelerating_residual_connection}

As shown in \cref{fig:proposed_methods} ~(b), Transformer blocks of DeiTs contain multiple residual connections requiring identical pruning indices. 
Both add layers of the Transformer block connect all the green-highlighted layers since the add layer serves as input for the last add layer in the Transformer block. 
In our analysis, all residual connections are linked by the last single residual connection “add\_24". 
Consequently, the importance scores of all filter indices are summed up to represent the importance of each filter index of “add\_24", and the filter with the least importance is removed to compress and accelerate DeiTs.

\section{Pruning ratios of SNP}

\begin{figure}[ht]
    \centering
    \includegraphics[width=\columnwidth]{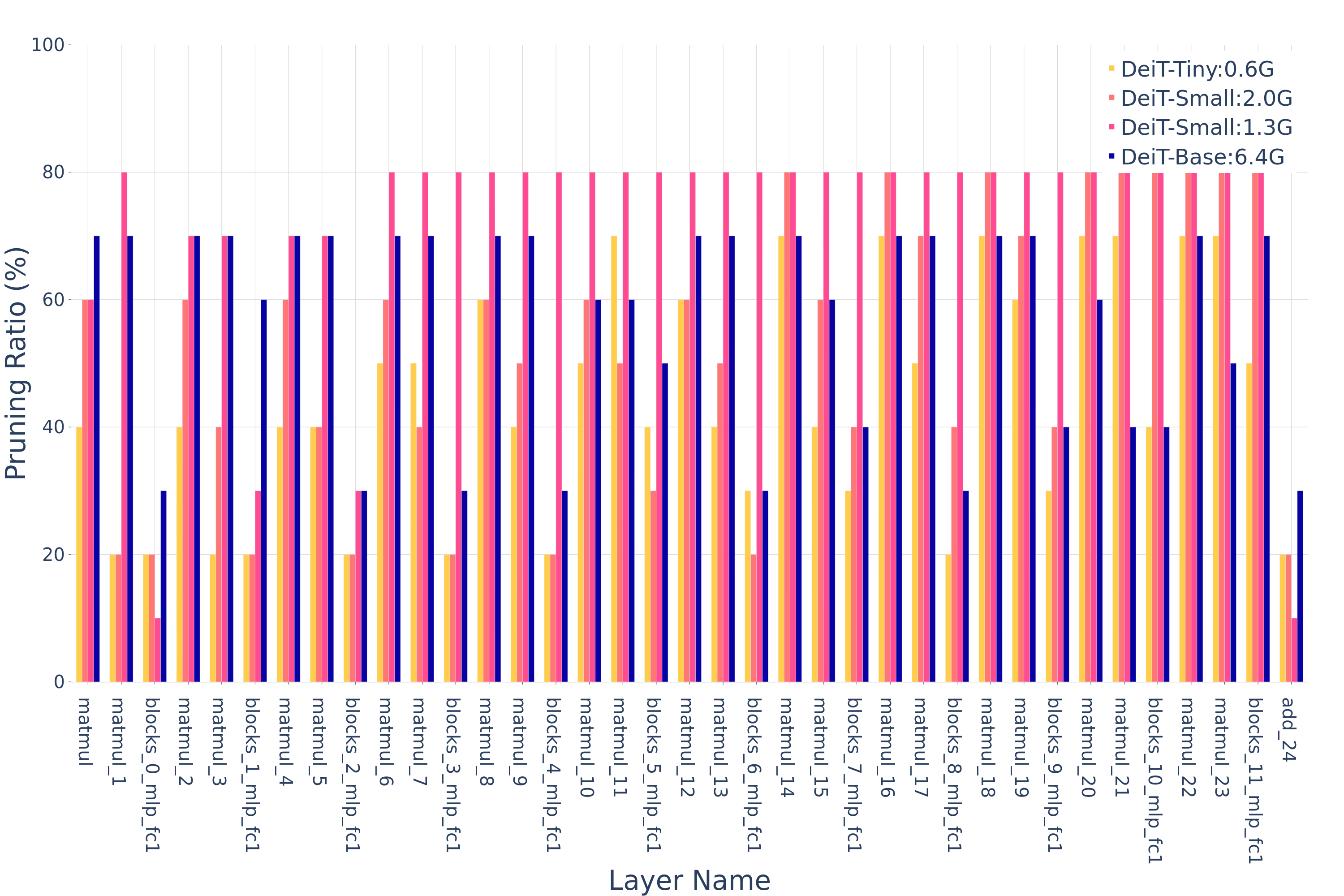}
    \caption{\textbf{Pruning ratios of all the compressed models using SNP.} 
    Even and odd numbers following the layer name “matmul" represent the matrix multiplication layers for the query and key layers, and the value layers, respectively.
    “add\_24" denotes the pruning ratios of all the connected layers by “add\_24".
    Each bar of pruning ratio is depicted as follows: DeiT-Tiny with SNP (0.6 GFLOPs), DeiT-Small (2.0 and 1.3 GFLOPs), and DeiT-Base (6.4 GFLOPs) respectively.
    }
    \label{fig:pruning_ratios}
\end{figure}

SNP is a graph-aware neuron-level pruning method designed for Transformer models. 
It is crucial to emphasize that our approach does not endorse a global pruning method to determine the optimal pruning ratio for each layer. 
As depicted in \cref{fig:pruning_ratios}, we provide layer-wise pruning ratios for the application of SNP to DeiTs. 
Even and odd numbers following the layer name “matmul" signify the matrix multiplication layers for the query and key layers, and the value layers, respectively. 
The layer named “add\_24" covers all residual connections, including those highlighted in green layers in ~\cref{fig:proposed_methods} ~(b) as mentioned in \cref{sec:sup_accelerating_residual_connection}.

In \cref{fig:pruning_ratios}, SNP removes over 59\% of the neurons in the MSA module, 34\% in the FFN, and 20\% across all interconnected layers via the residual connection layer, with only a 1.60\% performance degradation. 
Specifically, SNP removes more than 64\% on query and key layers, and 55\% on value layers. 
This indicates that DeiTs can maintain high performance even with a high pruning ratio in the MSA module, particularly in the query and key layers.

\section{Uniform pruning with SNP}

SNP introduces two pruning criteria for the MSA module, detailed in ~\cref{sec:preserving_attention_scores} and \cref{sec:value_layer}: preserving attention scores in query and key layers and eliminating inter-head redundancy in value layers.

\begin{figure}[ht]
    \centering
    \includegraphics[width=\columnwidth]{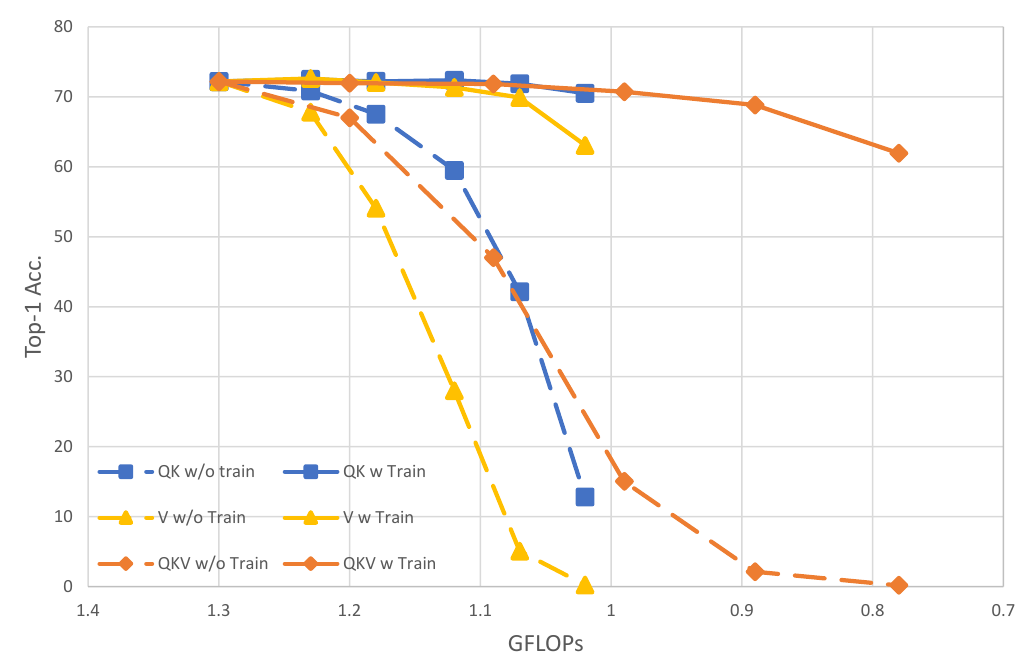}
    \caption{\textbf{Performance of the locally pruned model using SNP.} 
    Dashed line is the performance right after the model compression, while solid line represents the performance of the fine-tuned model.
    }
    \label{fig:local_prune}
\end{figure}

In ~\cref{fig:local_prune}, dashed lines show compressed model performance at specified pruning ratios (0\%, 10\%, 30\%, 50\%, 70\%, and 90\%), while solid lines depict fine-tuning performance after 30 epochs for each compressed model. 
The first method, preserving attention scores on query and key layers, is in blue, the second method, eliminating inter-head redundancy, is in yellow, and the combination is in orange.

Locally compressed models in ~\cref{fig:local_prune}, with distinct pruning criteria, exhibit identical FLOPs. Filter removal in query and key layers has no impact on other layers, while removal in the value layer affects subsequent layers, resulting in equivalent computational costs.

SNP on value layers maintains comparable performance with a 50\% filter removal (72.20\% to 71.35\%). 
SNP on query and key layers outperforms the original model by 0.14\%, even with 50\% of filters removed. It also sustains performance with only a 1.68\% drop (72.20\% to 70.52\%) when 90\% of query and key layers are removed.

These results highlight that not only can head pruning accelerate and compress Transformer models, but neuron-level pruning can achieve high compression ratios with minimal performance degradation.

\end{document}